\documentclass[conference]{IEEEtran}
\IEEEoverridecommandlockouts
\usepackage{cite}
\usepackage{amsmath,amssymb,amsfonts}
\usepackage{algorithmic}
\usepackage{algorithm}
\usepackage{caption}
\usepackage{subcaption}
\usepackage{booktabs}
\usepackage{textcomp}
\usepackage{cellspace}
\usepackage{hyperref}
\usepackage{graphicx}
\usepackage{textcomp}
\usepackage[T1]{fontenc}
\newcommand{\etal}{\emph{et al.}}
\newcommand{\hyperfootnote}[1][]{\def\ArgI\hyperfootnoteRelay}
\newcommand{\nofootnote}[1]{%
  \begingroup\def\thefootnote{}\footnotetext{#1}\endgroup}

\usepackage{xcolor}
\def\BibTeX{{\rm B\kern-.05em{\sc i\kern-.025em b}\kern-.08em
    T\kern-.1667em\lower.7ex\hbox{E}\kern-.125emX}}
\begin{document}

\title{Interpretability for Multimodal Emotion Recognition using Concept Activation Vectors}
\author{\IEEEauthorblockN{Ashish Ramayee Asokan}
\IEEEauthorblockA{\textit{Dept. of Computer Science} \\
\textit{PES University}\\
Bangalore, India \\
ashish.ramayee@gmail.com}
\and
\IEEEauthorblockN{Nidarshan Kumar}
\IEEEauthorblockA{\textit{Dept. of Computer Science} \\
\textit{PES University}\\
Bangalore, India\\
suryanidarshan@gmail.com}
\and
\IEEEauthorblockN{Anirudh V Ragam}
\IEEEauthorblockA{\textit{Dept. of Computer Science} \\
\textit{PES University}\\
Bangalore, India \\
anirudhragam19@gmail.com}
\and
\IEEEauthorblockN{Shylaja S S}
\IEEEauthorblockA{\textit{Dept. of Computer Science} \\
\textit{PES University}\\
Bangalore, India \\
shylaja.sharath@pes.edu}}

\maketitle

\begin{abstract}
Multimodal Emotion Recognition refers to the classification of input video sequences into emotion labels based on multiple input modalities (usually video, audio and text). In recent years, Deep Neural networks have shown remarkable performance in recognizing human emotions, and are on par with human-level performance on this task. Despite the recent advancements in this field, emotion recognition systems are yet to be accepted for real world setups due to the obscure nature of their reasoning and decision-making process. Most of the research in this field deals with novel architectures to improve the performance for this task, with a few attempts at providing explanations for these models’ decisions. In this paper, we address the issue of interpretability for neural networks in the context of emotion recognition using Concept Activation Vectors (CAVs). To analyse the model's latent space, we define human-understandable concepts specific to Emotion AI and map them to the widely-used IEMOCAP multimodal database. We then evaluate the influence of our proposed concepts at multiple layers of the Bi-directional Contextual LSTM (BC-LSTM) network to show that the reasoning process of neural networks for emotion recognition can be represented using human-understandable concepts. Finally, we perform hypothesis testing on our proposed concepts to show that they are significant for interpretability of this task.
\end{abstract}


\begin{IEEEkeywords}
Multimodal Emotion Recognition, Interpretability, Concept Activation Vectors, Emotion AI
\end{IEEEkeywords}

\section{Introduction}
\nofootnote{This work has been submitted to the IEEE for possible publication. Copyright may be transferred without notice, after which this version may no longer be accessible.}
The research of Machine Learning (ML) systems has witnessed a rapid growth in recent years, with their presence in diverse fields ranging from day-to-day use cases such as personal assistants and search engines to highly regulated domains involving high-risk decision-making such as medical diagnosis and autonomous driving. The increasing availability of large databases and hardware resources to train such complex ML systems have resulted in state-of-the-art performance across a wide range of tasks. However, despite these advancements, ML systems still lack transparency, i.e, the internal reasoning process of these models are hidden from the user, which can prove to be a pitfall that prevents humans from verifying the decisions made by these black box models \cite{carvalho2019machine}. Došilović \etal \cite{dovsilovic2018explainable} highlight the fact that Deep Neural Networks (DNNs) are criticized for serving only as approximations of a decision-making system whose decisions cannot be trusted. Therefore, these black box models must satisfy several assurances such as justifiability, usability, reliability, etc., for a practicable ML system. \par

Interpretability for Machine Learning can be defined as the extent to which a model's decisions can be consistently predicted or accounted for \cite{molnar2019}. According to Carvalho \etal, the taxonomy of interpretability methods is based on - (i) when the methods are applicable (\emph{Pre-Model}, \emph{In-Model}, \emph{Post-Model} (ii) whether the model is trained with a complexity constraint or analysed post-training (\emph{Intrinsic} vs \emph{Post hoc}) (iii) whether the interpretation is based on the model architecture (\emph{Model-Specific} vs \emph{Model-Agnostic}). There is often a trade-off between model complexity and model interpretability, i.e, the more complex a model is, the harder it is to interpret the decisions made by the same. This is especially the case with \emph{Intrinsic} methods where the model is trained with an additional complexity constraint to ensure effective interpretability, which affects model performance. However, \emph{Post Hoc} and \emph{Post Model} methods provide interpretability post-training, thereby ensuring no loss in performance.

With a better understanding of human emotions and the increasing availability of large emotion databases, emotion recognition has become an emerging research area in recent years. Emotions can be defined as a psycho-physiological process that is initiated by interaction with (or perception of) people or situations, with varying motivation and mood \cite{tripathi2018multi}. Emotion Recognition can be done using various modalities such as speech, text, EEG signals and facial expressions, among which facial expressions are more widely adopted due to easier availability of these datasets. Even though a large amount of work has been done in this field, emotion recognition is often challenging due to intra-class variance, i.e, variations in emotions among different ethnicities, cultures and age groups. In practice, it is observed that multimodal approaches are more robust to intra-class variance and often adopted by clinicians and psychologists. \cite{zhang2019context}. \par

Multimodal Emotion Recognition finds its application in the healthcare industry to provide a preliminary assessment of a patient's emotional state. Such systems have been used in a clinical setting for the diagnosis of medical conditions such as Schizophrenia and Autism \cite{grabowski2019emotional}. Due to the limited exploration of interpretability for emotion recognition by prior work, we address this problem using Concept Activation Vectors (CAVs) to determine which concepts a model uses to recognize human emotions. Based on cues used by clinicians to recognize emotions, we define appropriate concepts with the publicly available IEMOCAP multimodal database and evaluate their significance on the Bi-Directional Contextual LSTM network. In summary, our contributions are as follows:

\begin{itemize}
    \item We extend the existing \emph{Testing with Concept Activation Vectors (TCAV)} method to video, audio and text input, which is yet to be explored.
    
    \item We propose novel human-understandable concepts for interpreting multimodal emotion recognition and evaluate the significance of the same.
\end{itemize}

\section{Related Works}

This section provides an overview of the recent literature on Interpretable AI and Emotion Recognition. 

\subsection{Interpretable AI}

Interpretability aims to explain the reasoning process of DNNs through human-understandable terms to facilitate robustness and impartiality in decision-making. In addition to the broad classification of interpretability methods outlined in \emph{Sec. I}, the sub-classes of methods also include \emph{Feature Attribution Methods}, and \emph{Concept-based Methods}, discussed in detail below.

\subsubsection{Feature Attribution Methods} Feature attribution methods attempt to explain each individual prediction by determining the effect (positive or negative) of each input feature on the prediction. Local Interpretable Model-Agnostic Explanations (LIME) \cite{ribeiro2016should} and SHapley Additive exPlanations (SHAP) \cite{lundberg2017unified} are some of the most well-known general feature attribution methods. LIME attempts to construct interpretable classifiers on a perturbed dataset to interpret a given model, and SHAP proposes a method to compute an additive feature attribution score with desirable properties. A special case of feature attribution is \emph{Pixel Attribution (Saliency Maps)} that highlights relevant pixels for each individual prediction in image classification. Few of the methods discussed here are Grad-CAM, SmoothGrad and Integrated Gradients. Grad-CAM \cite{selvaraju2017grad} highlights the important regions of an input image for an individual prediction using the gradients of the final convolutional layer to generate an activation map. SmoothGrad \cite{smilkov2017smoothgrad} attempts to improve the visual quality of gradient-based sensitivity maps by averaging those of noisy versions of the input image. Integrated Gradients \cite{sundararajan2017axiomatic} provides pixel-level attribution by computing the path integral of the gradients between a baseline input and the regular input.


\subsubsection{Concept-based methods}

Concept-based methods aim to address interpretability in DNNs by extracting human-understandable concepts from a model's latent representations. Liu \etal \cite{liu2020explaining} propose a model distillation method based on unsupervised clustering that produces an \emph{Intrinsic} (interpretable by design) surrogate model. Kim \etal \cite{kim2018interpretability} introduce Concept Activation Vectors (CAVs) that use directional derivatives to represent human-understandable concepts from a model's activations and quantify the influence of a concept on the predictions of a single target class. Pfau \etal \cite{pfau2021robust} build on TCAV by providing global and local conceptual sensitivities and accounting for the non-linear influence of concepts on a model's predictions. Lucieri \etal \cite{lucieri2020interpretability} explore TCAV in the context of skin lesions classification using an InceptionV4 model built by the REasoning for COmplex Data (RECOD) Lab. Ghorbani \etal \cite{ghorbani2019towards} propose Automatic Concept-based Explanations (ACE) - a novel method that uses image segmentation and clustering to extract visual concepts used by a model.

\subsection{Emotion AI}

Emotion AI deals with the \emph{detection} and \emph{interpretation} of emotive channels involved in human communication. A considerable portion of Emotion AI research in recent years has dealt with performance improvements on the emotion recognition task through novel DNN architectures. Tripathi \etal \cite{tripathi2018multi} explore multimodal emotion recognition on the IEMOCAP database using speech, text and motion capture features. Mittal \etal \cite{mittal2020m3er} propose a novel fusion method to combine the face, text and speech modalities that is impervious to noise. Krishna \etal \cite{krishna2020multimodal} propose a cross-modal attention mechanism that uses audio and text features for emotion recognition. Majumder \etal \cite{majumder2018multimodal} and Poria \etal \cite{poria2017context} explore hierarchical contextual feature extraction for emotion recognition, which we adopt in this work. A comprehensive list of the architectures for Multimodal Emotion Recognition (MER) and Emotion Recognition in Conversation (ERC) is provided here (ERC,\footnote{\href{https://paperswithcode.com/sota/emotion-recognition-in-conversation-on}{IEMOCAP ERC Benchmark}}MER\footnote{\href{https://paperswithcode.com/sota/multimodal-emotion-recognition-on-iemocap}{IEMOCAP MER Benchmark}}), but our main focus is the former task.\par

Interpretability for multimodal emotion recognition has been explored with intrinsic and post-hoc methods primarily through EEG signals and speech input. Quing \etal \cite{qing2019interpretable} explore interpretable EEG-based emotion recognition using Emotional Activation Curves and evaluate their results on the DEAP and SEED dataset. Liu \etal \cite{liu2022group} propose Gated Bi-directional Alignment Network that effectively captures speech-text relations, and an interpretable Group Gated Fusion (GGF) layer that determines the significance of each modality through contribution weights. Mayou \etal \cite{mayor2021interpretable} propose a SincNet-based network for emotion classification with EEG signals that is interpreted by inspecting the filters learned by the model. Nguyen \etal \cite{nguyen2019multimodal} introduce a novel DNN architecture for multimodal emotion recognition and use non-linear Gaussian Additive Models to interpret the same. \par

A thorough survey of relevant literature showed that concept-based interpretation of multimodal emotion recognition is yet to be explored, and we attempt to address this gap by extending Concept Activation Vectors \cite{kim2018interpretability} to video, audio and textual data. We first define human-understandable concepts specific for emotion recognition based on inferences and observations from \cite{grabowski2019emotional}. The CAVs are then fitted to the BC-LSTM model's latent space at the chosen layers to compute the concept sensitivities and TCAV scores for each concept and for each target emotion.

\section{Proposed Methodology}

In this section, we discuss the feature extraction method used for multimodal emotion recognition and introduce our human-understandable concepts for interpreting DNNs with Concept Activation Vectors.

\begin{figure*}[t]
    \centering
    \begin{subfigure}[b]{\textwidth}
         \centering
         \includegraphics[width=0.7\textwidth]{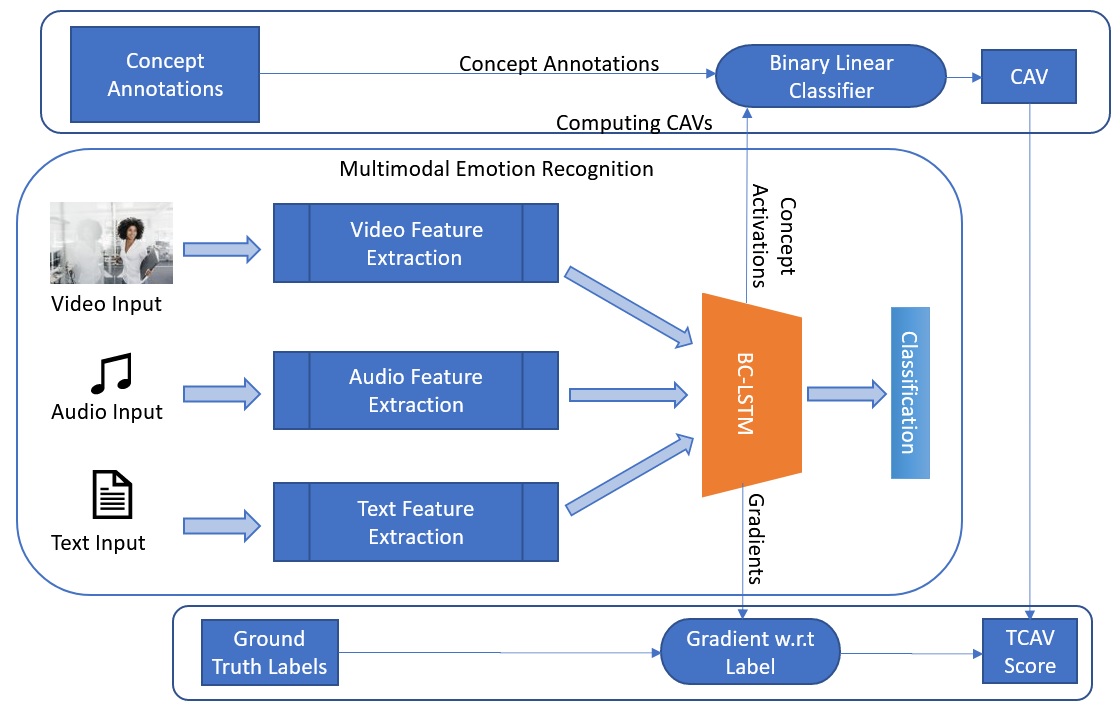}
         \caption{}
         \label{fig:a}
    \end{subfigure}
    \begin{subfigure}[b]{\textwidth}
    \centering
         \includegraphics[width=0.65\textwidth]{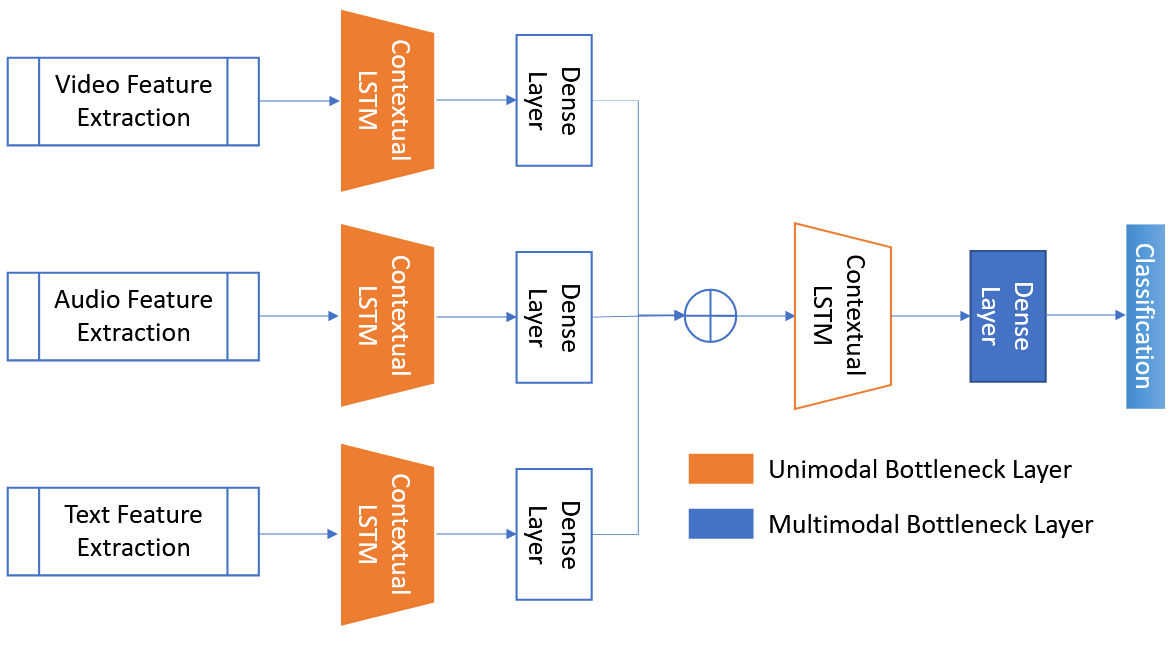}
         \caption{}
         \label{fig:a}
    \end{subfigure}
    \caption{(a) Overview of Methodology for TCAV on Emotion Recognition (b) Bottleneck layers chosen for TCAV on Emotion Recognition}
\end{figure*}

\subsection{Feature Extraction}

In this work, we adopt the feature extraction method described by Poria \etal \cite{poria2017context}. It is carried out in 2 stages: \emph{Context-Independent Extraction} that extracts the features for each input mode (audio, video and text) separately, and \emph{Context-dependent Extraction} that learns features across utterances for both unimodal and multimodal emotion recognition. \par

\subsubsection{Context-Independent Feature Extraction for Unimodal Input} Feature extraction on the unimodal input is done independent of the surrounding utterances and without any contextual information (or dependency). The steps involved in feature extraction for each input mode are discussed in detail below:

\begin{itemize}

    \item \textbf{Text Input.} The text inputs used for textual feature extraction are the transcripts for each of the utterances. As stated by Poria \etal, each of the utterances in a video are represented as a combination of 300 dimensional \emph{word2vec} vectors \cite{mikolov2013efficient} of each word in the utterances. The CNN used for feature extraction consists of 2 convolution layers and a single max-pool layer. The result of the pooling operation is projected onto a $d_t$-dimensional dense layer whose output serves as the input textual features for emotion recognition. \par
    
    \item \textbf{Video Input.} The feature extraction for visual input is done using a 3D-CNN, which is capable of learning features for each frame of the video as well as temporal features across video frames. The video input to the 3D-CNN has the dimensions ($c, h, w, f$), where $c$ is the number of channels (3 for RGB), $h$ and $w$ are the dimensions of each frame, and $f$ is the total number of frames. The 3D-CNN consists of a convolution layer with a 3D filter and a maxpool layer, followed by a dense layer of dimensions $d_v$. The output of this 3D-CNN is a $d_v$-dimensional vector that represents the utterance-level input visual features. \par
    
    \item \textbf{Audio Input.} Audio feature extraction is done using the openSMILE open-source software \cite{eyben2010opensmile} that automatically extracts essential audio features. OpenSMILE extracts several low-level features such as pitch, intensity, MFCC, etc. These features serve as the input audio features for the emotion recognition model.
\end{itemize}

\subsubsection{Context-Dependent Feature Extraction}

The contextual features are extracted using the \emph{Contextual-LSTM} architecture proposed by Poria \etal, which is a part of their Bi-directional Contextual LSTM network (\emph{Fig. 1b}). The intuition behind this architecture is that the surrounding utterances can provide essential information in the classification of the current utterance, thereby requiring a model that takes such dependencies into consideration. Let $X_k$ represent the input features for utterance $k$ and $L_k$ represent the output of the LSTM network for utterance $k$. The output for the next utterance $L_{k+1}$ depends on $X_{k+1}$ as well as the output of the previous LSTM network $L_{k}$, which represents the learning of contextual information. This contextual-LSTM module is used for unimodal and multimodal feature extraction.

\subsection{Interpretability using CAVs}

\subsubsection{Testing with Concept Activation Vectors (TCAV)}
To achieve interpretability in terms of human-understandable concepts, Kim \etal \cite{kim2018interpretability} proposed Concept Activation Vectors (CAVs) - a linear interpretability method that represents a concept with a vector in a neural network's activation space given a set of positive and negative examples representing the concept. Given a positive examples set $X^{+ve}$ and a negative examples set $X^{-ve}$, a binary classifier $v_C^l$ is trained to distinguish between the activations of the positive examples set $f_l(x), x \in X^{+ve}$ and the negative examples set $f_l(x), x \in X^{-ve}$, where $f_l(x)$ represents the neural activation at a layer $l$ of a network. This binary classifier $v_C^l$ represents the CAV for the concept $C$ at layer $l$. The \emph{Testing with CAV (TCAV)} method introduces a metric known as the TCAV score that represents a ML model's sensitivity to a particular concept across all class labels. Given a concept $C$, the TCAV score at a layer $l$ for examples belonging to class $k$ ($X_k)$ is given as:

\begin{align}
    TCAV_{Q_{C, k, l}} = \frac{|x \in X_k : S_{C, l, k}(x) > 0|}{{X_k}}
\end{align}

where $S_{C, l, k}$ represents the directional derivative at layer $l$ for concept $C$ and class $k$ given by $S_{C, l, k} = f_l'(x) . v_C^l$ ($f_l'(x)$ is the derivative of the activation at layer $l$). TCAV provides a quantitative measure of conceptual sensitivity across entire input classes and can be extended to input modes other than images.

\subsubsection{TCAV for Emotion Recognition}

Here, we delineate the concepts used to interpret multimodal emotion recognition models with TCAV. We define a single concept for each of video, audio and text input modes - \emph{Variations in Physiognomy}, \emph{Voice Pitch} and \emph{Utterance Polarity}, which are discussed below:

\begin{itemize}
\item \textbf{Variations in Physiognomy (Deviation from Neutral Expressions).} Emotions such as anger and excitement capture extreme changes in facial expressions compared to the neutral resting face. They are characterized by changes in the facial features such as eye contact, lip movement, etc. According to Grabowski \etal \cite{grabowski2019emotional}, analysis of emotions in a valence/arousal spectrum allows for the distinction between neutral and extreme emotions. Specifically, emotions associated with high arousal are characterized by this concept. Positive examples are those utterances that show extreme variations in facial expressions and negative examples represent the neutral resting face.

\item \textbf{Voice Pitch.}  Among the several sound features responsible for emotional prosody, pitch is one of the features essential for emotion recognition, which is defined as the relative highness or lowness of tone perceived by the human ear. Quinto \etal \cite{quinto2013emotional} hypothesise that high and low pitch are associated with specific emotions in the speech domain. For instance, anger and excitement are associated with a high pitch while sadness is associated with low pitch. In our work, we assume that anger, frustration and excitement are often associated with high pitch and thereby serve as positive examples for this concept, while negative examples have relatively lower pitch indicating emotions such as sadness or neutral.

\item \textbf{Utterance Polarity.} Since the text inputs used for emotion recognition are the transcripts for the utterances, we use the underlying sentiment (positive or negative) as a concept for interpretability. Each utterance from the input is assigned a polarity score from -1 to 1. Utterances associated with positive emotions such as happiness and excitement have positive polarity and utterances associated with negative emotions have negative utterance polarity.
\end{itemize}

\begin{algorithm}
 \caption{TCAV for Emotion Recognition}
 \begin{algorithmic}[1]
 \renewcommand{\algorithmicrequire}{\textbf{Input:}}
 \renewcommand{\algorithmicensure}{\textbf{Output:}}
 \REQUIRE Layer $l$, $C = \{C_1, C_2, \dots C_n\}$
 \REQUIRE $D = \{(X^{C_1}, Y^{C_1}), (X^{C_2}, Y^{C_2}), \dots (X^{C_n}, Y^{C_n})\}$

  \FOR {$i = 1$ to $n$}
    \STATE $D_i = (f_l(X^{C_i}), Y^{C_i})$ \COMMENT{data for concept $i$}
    \STATE $D_i^U = VideoToUtterance(D_i)$ 
    \STATE $v^{C_i}_l = BinaryClassifier(D_i^U)$ \COMMENT{CAV}
    \item[]
    \FOR {$k = 1$ to $K$}
        \STATE $h = VideoToUtterance(X^{C_i}_k)$
        \STATE $S_{C,k,l} = f'_l(h) \times v^{C_i}_l$
        \STATE $TCAV_{Q_{C, k, l}} = TCAVScore(h, S_{C,k,l})$
    \ENDFOR
\ENDFOR
\end{algorithmic}
\end{algorithm}

The sequence of steps involved in computing conceptual sensitivities for emotion recognition are outlined in \emph{Algo. 1}. Given the set of concepts $C$ and the concept annotations set $D$, we wish to compute the CAV for each concept along with the TCAV scores for each concept $C_i$, label $k$ and layer $l$. Similar to the original TCAV method, the activations of the concept examples at layer $l$ are extracted from the network and a binary classifier is fitted to these activations. However, there is an additional step while extending TCAV to emotion recognition, which is the conversion of video-level activations to utterance-level activations (Steps 3,6 in \emph{Algo. 1}). The raw activations have the dimensions ($n, t, f$), where $n$ is the number of videos, $t$ is the sequence length and $f$ is the number of features. These activations are reshaped to ($n \times t, f$) so that the first dimension represents the number of utterances. Therefore, the reshaped activations represent the utterance-level activations of the model at layer $l$. A binary classifier is then fitted to the concepts sets for each concept $C_i$ to obtain the CAV $v_l^{C_i}$. The conceptual sensitivities and TCAV scores are computed as explained in \emph{Sec. III.B.1}.

\section{Experiments and Results}

In this section, we cover the experimental setup used for training the multimodal emotion recognition model and interpreting the same using Concept Activation Vectors. 

\subsection{Model}

To determine the influence of our concepts for emotion recognition, we make use of the Bi-directional Contextual LSTM network (\emph{Fig. 1b}) introduced by Poria \etal \cite{poria2017context} trained on the IEMOCAP multimodal database (\emph{Sec. IV.B}). The motivation behind choosing this architecture is the fact that this is one of the few simple and straightforward speaker-independent multimodal architectures for emotion recognition, which makes interpreting its decisions more convenient. The current state-of-the-art methods \cite{yang2021hybrid} \cite{kim2021emoberta} for emotion recognition (in conversation) on IEMOCAP make use of speaker-specific components to enhance performance, which is outside the scope of our work. Contextual Hierarchical Fusion \cite{majumder2018multimodal} extends the idea of contextual information to 3 hierarchical levels but provides only a marginal improvement over BC-LSTM, thereby making BC-LSTM the appropriate choice for our work. Bi-directional LSTMs are used here to account for contextual information from the preceding and following utterances for emotion classification. Fusion of the modalities is done in a hierarchical fashion consisting of 2 levels. \emph{Level 1} extracts context-sensitive information from the context-independent features that are fed to the contextual-LSTM module. These context-sensitive unimodal features are then concatenated and fed to the final contextual-LSTM module to extract context-sensitive multimodal features. For all 6 emotion labels of the IEMOCAP database, the BC-LSTM network achieves $41.7\%$ on video input, $47.4\%$ on audio input, $53.7\%$ on text input and $57.5\%$ with all 3 inputs combined, which is in accordance to the results presented in PapersWithCode \footnote{\href{https://paperswithcode.com/sota/multimodal-emotion-recognition-on-iemocap}{PapersWithCode - IEMOCAP Benchmark}}.

\begin{figure*}[t]
    \centering
    \begin{subfigure}{0.4\textwidth}
        \includegraphics[width=\textwidth]{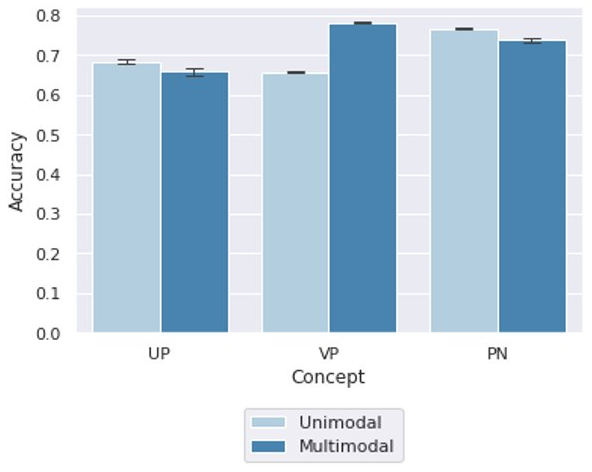}
        \caption{}
        \label{fig:1}
    \end{subfigure}
    \hspace{1cm}
    \begin{subfigure}{0.4\textwidth}
        \includegraphics[width=\textwidth]{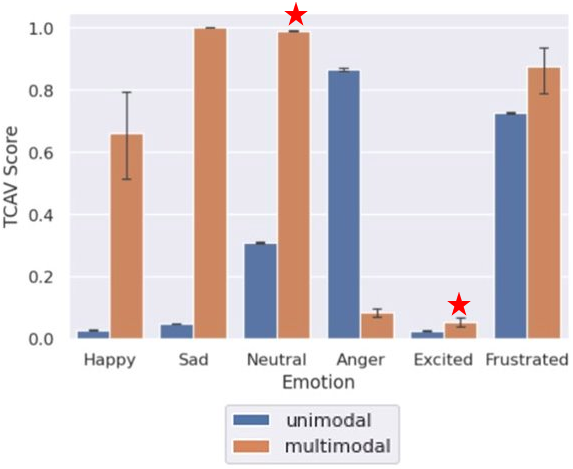}
        \caption{}
        \label{fig:2}
    \end{subfigure}
    \hfill
    \begin{subfigure}{0.4\textwidth}
        \includegraphics[width=\textwidth]{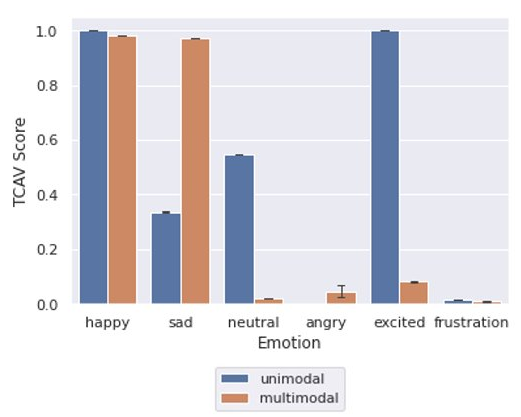}
        \caption{}
        \label{fig:3}
    \end{subfigure}
    \hspace{1cm}
    \begin{subfigure}{0.4\textwidth}
        \includegraphics[width=\textwidth]{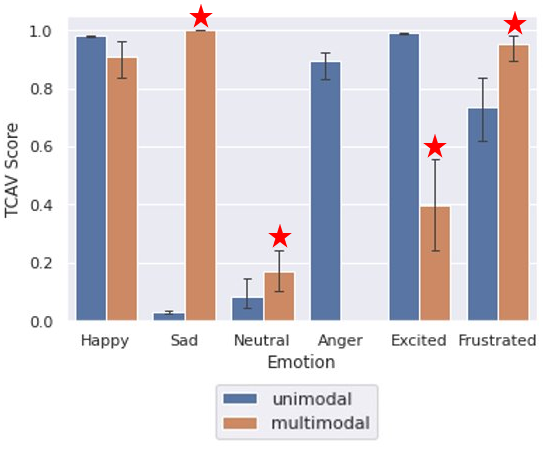}
        \caption{}
        \label{fig:4}
    \end{subfigure}
    \caption{(a) Unimodal \& Multimodal CAV accuracy for all 3 concepts. (b) TCAV scores for \emph{PT} concept. (c) TCAV scores for \emph{VP} ceoncept. (d) TCAV scores for \emph{UP} concept (\emph{Stars indicate insignificant concepts)}}
    \label{fig:figures}
\end{figure*}

\subsection{Dataset}

The dataset used to train the BC-LSTM network and define human-understandable concepts for TCAV is the Interactive Emotional Dyadic Motion Capture (IEMOCAP) database \cite{busso2008iemocap}. It consists of scripted acts and improvisations involving 10 speakers. Each video involves a conversation between 2 subjects divided into several utterances and each of these utterances is associated with one of the following 6 emotion labels: happy ($0$), sad ($1$), neutral ($2$), angry ($3$), excited ($4$) and frustrated ($5$). To train the BC-LSTM network, we use a 70-30 split for the training and testing sets, i.e, the training set contains 121 videos (4290 utterances) and the testing set contains 31 videos (1208 utterances).

\subsection{Experimental Setup}

All the experiments are carried out on the BC-LSTM network with concept examples from the IEMOCAP database. The hierarchical components of the BC-LSTM architecture are trained sequentially and separately, i.e, the model is not trained in an "\emph{end-to-end}" fashion. The unimodal contextual-LSTM modules are trained separately and frozen while training the multimodal contextual-LSTM module. The unimodal contextual-LSTM modules are trained separately with the Adam optimizer for $100$ epochs at a learning rate of $1 \times 10^{-4}$. Based on trials with all the layers of the model, we found that extracting the activations from the \emph{contextual-LSTM} layer at the unimodal level and the \emph{Dense} layer at the multimodal level gave the best results. The samples distribution for the 3 concepts are outlined in \emph{Table 1}. The concept examples are selected from the top $n$ videos that have the maximum number of utterances from the emotion labels for positive and negative examples indicated in \emph{Table 1}. \par

\begin{table}[h!]
    \centering
    \begin{tabular}{Sc Sc Sc Sc Sc} 
    \specialrule{1pt}{1pt}{1pt}
    
    \textbf{Concept} &
    \multicolumn{2}{c}{\textbf{No. of Samples}} &
    \multicolumn{2}{c}{\textbf{Labels of Samples}} \\
    & \textbf{+ve} & \textbf{-ve} & \textbf{\emph{+ve}} & \textbf{\emph{-ve}} \\
    \specialrule{1pt}{1pt}{1pt}
    \begin{tabular}{@{}c@{}}Variations in \\ Physiognomy (VP)\end{tabular} & 2200 & 2200 & 0,4,5 & 2 \\
    \hline
    \begin{tabular}{@{}c@{}}Utterance \\ Polarity (UP)\end{tabular} & 1361 & 792 & 0,2,4 & 1,3,5 \\
    \hline
    \begin{tabular}{@{}c@{}}Voice \\ Pitch (PT)\end{tabular} & 620 & 1706 & 0\,4\,5 & 2 \\
    \hline
    \vspace{0.5cm}
    \end{tabular}
    \caption{Sample distribution for our proposed concepts}
\end{table}

Examples for the \emph{VP} concept are collected solely based on the assumption from \emph{Sec III.B}. The positive examples set consists of preprocessed utterances belonging to the \emph{happy}, \emph{excited} and \emph{frustrated} emotion classes based on manual inspection of a small subset of videos from the IEMOCAP database. For the \emph{PT} concept, we use Self-supervised Pitch Estimation proposed by Gfeller \etal \cite{gfeller2020spice} to estimate the pitch for every utterance from the concept set and assign positive/negative labels based on a threshold pitch value of $250Hz$, i.e, the preprocessed utterance belongs to the positive examples set for \emph{PT} if the pitch of the utterance exceeds $250Hz$. To compute the text polarity of utterances, we use the \emph{TextBlob} Python library that assigns a text polarity of -1 to 1 for each utterance based on a weighted average sentiment score of the words in the utterance. \par

To account for variations in the binary classifiers' initialization and preprocessing of the concept examples \cite{lucieri2020interpretability}, the training of the CAVs is repeated 30 times resulting in 30 different vectors. We evaluate the statistical significance of our concepts by training 50 random CAVs for each layer and assigning random labels. We then perform a 2-tailed $t$-test on the TCAV score distributions of the random concepts and the proposed concepts at a significance level $\alpha=0.05$. \par

\subsection{Results}

Here, we discuss the quantitative evaluation of the CAVs for our proposed concepts through the classifier accuracies, TCAV scores and hypothesis tests for concept significance. Since there is no quantitative method to compare interpretability methods and due to the lack of results for concept-based interpretation of emotion recognition, we evaluate our concepts without any comparison to prior work. \par

\emph{Fig. 2a} shows the test accuracies of the classifiers for the CAVs at the unimodal and multimodal levels of the BC-LSTM network trained on the IEMOCAP multimodal database. The overall accuracies are relatively low due to the fact that linear classifiers are used to define the concepts. The graph also shows that there is minimal variation in the classifiers' accuracies, indicating that the 30 different vectors are consistent for each of the concepts.

\subsubsection{Variations in Physiognomy (VP)}

The examples used to represent \emph{VP} are collected based on the general assumption that emotions showing medium to high arousal such as excitement, joy, and anger \cite{grabowski2019emotional} display a greater deviation from the neutral resting face and can be distinctly identified through variations in visual features. \emph{Fig. 2b} indicates the TCAV scores for \emph{VP} at the unimodal and multimodal bottleneck layers. We see that the highest scores are observed for the \emph{happy} and \emph{excited} classes, while the scores for \emph{sad}, \emph{angry} and \emph{frustration} classes are relatively lower, which confirms the assumption stated above. However, the scores for \emph{frustrated} and \emph{neutral} classes are not in line with this assumption. At the multimodal bottleneck, it is observed that \emph{VP} has a strong influence on \emph{happy} and \emph{sad} classes but much lower influence on the rest of the classes. The consistently strong influence of \emph{VP} on the \emph{happy} class across the unimodal and multimodal bottlenecks is evidence for the fact that \emph{VP} is essential for recognizing happiness. It is also observed that the scores for the \emph{sad}, \emph{excited} and \emph{frustration} classes are exceptions for the general assumption stated earlier. Since the concept set is not created based on a quantitative measure for variations in facial expressions, it is possible that the inconsistencies in TCAV scores are due to the nature of the concept set given that it is only an approximation based on the general assumption. Another factor that could contribute to this discrepancy is the relatively poor performance of the BC-LSTM network on video input from IEMOCAP as mentioned in \emph{Sec. IV.A}. \par

\subsubsection{Voice Pitch (PT)}

Pitch of an individual's voice can be used as a strong indicator of expression of specific emotions. \emph{Fig. 2b} shows the TCAV scores for \emph{PT} at the unimodal and multimodal bottlenecks. It is observed that \emph{PT} has the highest influence (0.865 and 0.726) on the \emph{anger} and \emph{frustration} classes at the unimodal bottleneck. This observation is consistent with \cite{quinto2013emotional}, in that expression of such emotions are associated with high pitch and \emph{PT} can be used as a distinguishing trait. This, however, is not true for the \emph{excitement} class, which ideally is characterized by high pitch. At the multimodal level, it is observed that the \emph{happy}, \emph{sad}, \emph{neutral}, and \emph{frustrated} emotion classes have high scores for \emph{PT} ($0.843$, $1$, $0.988$ and $0.875$ respectively). This is a contradiction to the general presumption that only emotions with high arousal are associated with high pitch and that pitch can be used as a distinguishing factor. This discrepancy could be due to the nature of the utterances found in the IEMOCAP database. It is observed that some of the utterances for emotions with medium arousal (\emph{happy}, \emph{sad}, etc.) have higher pitch than some of the utterances for emotions with high arousal.

\subsubsection{Utterance Polarity (UP)}

\emph{Fig. 2d} shows the TCAV scores for \emph{UP} at the unimodal and multimodal bottleneck layers. From the scores, it is evident that \emph{UP} has a high influence on all the target emotions except the \emph{sad} and \emph{neutral} labels at the unimodal bottleneck, which is as expected. Phrases that depict emotions involving medium to high arousal (intensity) tend to have a high level of sentiment polarity compared to the neutral emotion. We observe that the influence of \emph{UP} on the \emph{neutral} and \emph{sad} target emotions is relatively low, which is in line with common observations on emotion recognition using text. Emotions such as neutral and sadness are not as conveniently distinguishable as the other target emotions. Given a phrase from these emotion classes, the polarity is approximately $0$, which makes it difficult to differentiate utterances of the \emph{neutral} and \emph{sad} classes. At the multimodal bottleneck layer, the scores are negligible for the \emph{neutral} and \emph{angry} classes and significant for the \emph{happy} and \emph{anger} classes. This signifies that \emph{UP} is insignificant towards the classification of examples into the \emph{sad}, \emph{neutral}, \emph{excited} and \emph{frustrated} emotions labels and that \emph{VP}, \emph{PT} concepts play a more important role for these classes. Despite the high TCAV scores of \emph{UP} for the \emph{sad} and \emph{frustrated} classes, hypothesis tests shows insignificance of the concept for these labels. \par

\subsubsection{Hypothesis Testing}

To test the significance of the proposed concepts for emotion recognition, we perform a 2-tailed $t$-test. We first generate 50 random concept sets from the training set and assign positive and negative labels in a random fashion to the activations from the unimodal and multimodal bottleneck layers. This is followed by fitting binary classifiers to these random concept sets to generate 50 random CAVs. We then perform a hypothesis test by conducting a 2 tailed $t$-test for the distribution of the TCAV scores for the proposed concepts and the 50 random concepts at a significance level $\alpha=0.05$. The null and alternate hypotheses are defined as follows:

\begin{align}
    \textrm{Null Hypothesis } H_o: \mu_r = \mu_t \\
    \textrm{Alternate Hypothesis } H_a: \mu_r \neq \mu_t
\end{align}

    

Here, $\mu_r$ represents the mean score of the random concepts distribution and $\mu_t$ represents the mean score of the proposed concepts score distributions. We consider a concept to be significant if the null hypothesis is rejected for 40 of the 50 random TCAV score distributions. It is observed that (\emph{Fig. 2b, 2c, 2d}) the \emph{UP} concept is significant for all emotion classes at the unimodal bottleneck and insignificant for the \emph{sad}, \emph{neutral}, \emph{excited} and \emph{frustrated} classes. The \emph{VP} concept is significant for all emotion classes at both the unimodal and multimodal bottlenecks. The \emph{PT} concept is significant for all emotion classes at the unimodal level but insignificant for the \emph{neutral} and \emph{excited} emotions at the multimodal level.

\section{Conclusions and Future Work}

Emotion AI has been widely used in critical domains such as medical diagnosis and security, and interpretability for emotion recognition will ensure the robustness and reliability of affective computing systems. To this end, we explore concept-based interpretation of emotion recognition through Concept Activation Vectors (CAVs) to quantify the influence of emotion-related concepts for a typical multimodal emotion recognition model. We define novel concepts based on existing Emotion AI literature, and analyse the relevance of the same. Through our results, it is evident that DNNs for emotion recognition make use of human-understandable concepts for classification, just like humans. \par

We further evaluate the significance of our concepts through hypothesis testing on the TCAV scores. The results show that the multimodal architecture makes use of specific concepts for specific emotion classes. There is no single concept that is significant for all the emotions, which is in line with pre-established notions for this task from the human perspective. Current literature on this task shows that most of the models trained on the IEMOCAP database tend to perform better on text input than the other two input modes, which could affect the interpretation of these models. Thus, one of the focuses for future work can be the interpretation of emotion recognition models that are independent of dataset bias. \par

In this work, we have explored the interpretability for emotion recognition on the BC-LSTM network, which is relatively simple compared to the state-of-the-art models. The TCAV method can be extended to more complex architectures to evaluate our concepts for these models. In addition to these improvements, the discovery of emotion-related concepts in an unsupervised setting can be a possible direction for future research. This can reduce human effort in annotating concept examples for emotion classification and enhance the interpretability of DNNs by allowing models to generate their own concepts. \par

\bibliography{citations}

\begin{thebibliography}{10}

\bibitem{carvalho2019machine}
D.~V. Carvalho, E.~M. Pereira, and J.~S. Cardoso, ``Machine learning
  interpretability: A survey on methods and metrics,'' {\em Electronics},
  vol.~8, no.~8, p.~832, 2019.

\bibitem{dovsilovic2018explainable}
F.~K. Do{\v{s}}ilovi{\'c}, M.~Br{\v{c}}i{\'c}, and N.~Hlupi{\'c}, ``Explainable
  artificial intelligence: A survey,'' in {\em 2018 41st International
  convention on information and communication technology, electronics and
  microelectronics (MIPRO)}, pp.~0210--0215, IEEE, 2018.

\bibitem{molnar2019}
C.~Molnar, {\em Interpretable Machine Learning}.
\newblock 2019.

\bibitem{tripathi2018multi}
S.~Tripathi, S.~Tripathi, and H.~Beigi, ``Multi-modal emotion recognition on
  iemocap dataset using deep learning,'' {\em arXiv preprint arXiv:1804.05788},
  2018.

\bibitem{zhang2019context}
M.~Zhang, Y.~Liang, and H.~Ma, ``Context-aware affective graph reasoning for
  emotion recognition,'' in {\em 2019 IEEE International Conference on
  Multimedia and Expo (ICME)}, pp.~151--156, IEEE, 2019.

\bibitem{grabowski2019emotional}
K.~Grabowski, A.~Rynkiewicz, A.~Lassalle, S.~Baron-Cohen, B.~Schuller,
  N.~Cummins, A.~Baird, J.~Podg{\'o}rska-Bednarz, A.~Pieni{\k{a}}{\.z}ek, and
  I.~{\L}ucka, ``Emotional expression in psychiatric conditions: New technology
  for clinicians,'' {\em Psychiatry and clinical neurosciences}, vol.~73,
  no.~2, pp.~50--62, 2019.

\bibitem{ribeiro2016should}
M.~T. Ribeiro, S.~Singh, and C.~Guestrin, ``" why should i trust you?"
  explaining the predictions of any classifier,'' in {\em Proceedings of the
  22nd ACM SIGKDD international conference on knowledge discovery and data
  mining}, pp.~1135--1144, 2016.

\bibitem{lundberg2017unified}
S.~M. Lundberg and S.-I. Lee, ``A unified approach to interpreting model
  predictions,'' in {\em Proceedings of the 31st international conference on
  neural information processing systems}, pp.~4768--4777, 2017.

\bibitem{selvaraju2017grad}
R.~R. Selvaraju, M.~Cogswell, A.~Das, R.~Vedantam, D.~Parikh, and D.~Batra,
  ``Grad-cam: Visual explanations from deep networks via gradient-based
  localization,'' in {\em Proceedings of the IEEE international conference on
  computer vision}, pp.~618--626, 2017.

\bibitem{smilkov2017smoothgrad}
D.~Smilkov, N.~Thorat, B.~Kim, F.~Vi{\'e}gas, and M.~Wattenberg, ``Smoothgrad:
  removing noise by adding noise,'' {\em arXiv preprint arXiv:1706.03825},
  2017.

\bibitem{sundararajan2017axiomatic}
M.~Sundararajan, A.~Taly, and Q.~Yan, ``Axiomatic attribution for deep
  networks,'' in {\em International Conference on Machine Learning},
  pp.~3319--3328, PMLR, 2017.

\bibitem{liu2020explaining}
Y.-h. Liu and S.~O. Arik, ``Explaining deep neural networks using unsupervised
  clustering,'' {\em arXiv preprint arXiv:2007.07477}, 2020.

\bibitem{kim2018interpretability}
B.~Kim, M.~Wattenberg, J.~Gilmer, C.~Cai, J.~Wexler, F.~Viegas, {\em et~al.},
  ``Interpretability beyond feature attribution: Quantitative testing with
  concept activation vectors (tcav),'' in {\em International conference on
  machine learning}, pp.~2668--2677, PMLR, 2018.

\bibitem{pfau2021robust}
J.~Pfau, A.~T. Young, J.~Wei, M.~L. Wei, and M.~J. Keiser, ``Robust semantic
  interpretability: Revisiting concept activation vectors,'' {\em arXiv
  preprint arXiv:2104.02768}, 2021.

\bibitem{lucieri2020interpretability}
A.~Lucieri, M.~N. Bajwa, S.~A. Braun, M.~I. Malik, A.~Dengel, and S.~Ahmed,
  ``On interpretability of deep learning based skin lesion classifiers using
  concept activation vectors,'' in {\em 2020 International Joint Conference on
  Neural Networks (IJCNN)}, pp.~1--10, IEEE, 2020.

\bibitem{ghorbani2019towards}
A.~Ghorbani, J.~Wexler, J.~Zou, and B.~Kim, ``Towards automatic concept-based
  explanations,'' {\em arXiv preprint arXiv:1902.03129}, 2019.

\bibitem{mittal2020m3er}
T.~Mittal, U.~Bhattacharya, R.~Chandra, A.~Bera, and D.~Manocha, ``M3er:
  Multiplicative multimodal emotion recognition using facial, textual, and
  speech cues,'' in {\em Proceedings of the AAAI Conference on Artificial
  Intelligence}, vol.~34, pp.~1359--1367, 2020.

\bibitem{krishna2020multimodal}
D.~Krishna and A.~Patil, ``Multimodal emotion recognition using cross-modal
  attention and 1d convolutional neural networks.,'' in {\em Interspeech},
  pp.~4243--4247, 2020.

\bibitem{majumder2018multimodal}
N.~Majumder, D.~Hazarika, A.~Gelbukh, E.~Cambria, and S.~Poria, ``Multimodal
  sentiment analysis using hierarchical fusion with context modeling,'' {\em
  Knowledge-based systems}, vol.~161, pp.~124--133, 2018.

\bibitem{poria2017context}
S.~Poria, E.~Cambria, D.~Hazarika, N.~Majumder, A.~Zadeh, and L.-P. Morency,
  ``Context-dependent sentiment analysis in user-generated videos,'' in {\em
  Proceedings of the 55th annual meeting of the association for computational
  linguistics (volume 1: Long papers)}, pp.~873--883, 2017.

\bibitem{qing2019interpretable}
C.~Qing, R.~Qiao, X.~Xu, and Y.~Cheng, ``Interpretable emotion recognition
  using eeg signals,'' {\em Ieee Access}, vol.~7, pp.~94160--94170, 2019.

\bibitem{liu2022group}
P.~Liu, K.~Li, and H.~Meng, ``Group gated fusion on attention-based
  bidirectional alignment for multimodal emotion recognition,'' {\em arXiv
  preprint arXiv:2201.06309}, 2022.

\bibitem{mayor2021interpretable}
J.~M. Mayor-Torres, M.~Ravanelli, S.~E. Medina-DeVilliers, M.~D. Lerner, and
  G.~Riccardi, ``Interpretable sincnet-based deep learning for emotion
  recognition from eeg brain activity,'' in {\em 2021 43rd Annual International
  Conference of the IEEE Engineering in Medicine \& Biology Society (EMBC)},
  pp.~412--415, IEEE, 2021.

\bibitem{nguyen2019multimodal}
T.-L. Nguyen, S.~Kavuri, and M.~Lee, ``A multimodal convolutional neuro-fuzzy
  network for emotion understanding of movie clips,'' {\em Neural Networks},
  vol.~118, pp.~208--219, 2019.

\bibitem{mikolov2013efficient}
T.~Mikolov, K.~Chen, G.~Corrado, and J.~Dean, ``Efficient estimation of word
  representations in vector space,'' {\em arXiv preprint arXiv:1301.3781},
  2013.

\bibitem{eyben2010opensmile}
F.~Eyben, M.~W{\"o}llmer, and B.~Schuller, ``Opensmile: the munich versatile
  and fast open-source audio feature extractor,'' in {\em Proceedings of the
  18th ACM international conference on Multimedia}, pp.~1459--1462, 2010.

\bibitem{quinto2013emotional}
L.~R. Quinto, W.~F. Thompson, and F.~L. Keating, ``Emotional communication in
  speech and music: The role of melodic and rhythmic contrasts,'' {\em
  Frontiers in psychology}, vol.~4, p.~184, 2013.

\bibitem{yang2021hybrid}
L.~Yang, Y.~Shen, Y.~Mao, and L.~Cai, ``Hybrid curriculum learning for emotion
  recognition in conversation,'' {\em arXiv preprint arXiv:2112.11718}, 2021.

\bibitem{kim2021emoberta}
T.~Kim and P.~Vossen, ``Emoberta: Speaker-aware emotion recognition in
  conversation with roberta,'' {\em arXiv preprint arXiv:2108.12009}, 2021.

\bibitem{busso2008iemocap}
C.~Busso, M.~Bulut, C.-C. Lee, A.~Kazemzadeh, E.~Mower, S.~Kim, J.~N. Chang,
  S.~Lee, and S.~S. Narayanan, ``Iemocap: Interactive emotional dyadic motion
  capture database,'' {\em Language resources and evaluation}, vol.~42, no.~4,
  pp.~335--359, 2008.

\bibitem{gfeller2020spice}
B.~Gfeller, C.~Frank, D.~Roblek, M.~Sharifi, M.~Tagliasacchi, and
  M.~Velimirovi{\'c}, ``Spice: Self-supervised pitch estimation,'' {\em
  IEEE/ACM Transactions on Audio, Speech, and Language Processing}, vol.~28,
  pp.~1118--1128, 2020.

\end{thebibliography}
\bibliographystyle{ieeetr}

\end{document}